\documentclass[10pt,twocolumn]{article}

\PassOptionsToPackage{table,xcdraw}{xcolor}
\usepackage[preprint]{antintlpaper}
\usepackage{inconsolata}
\usepackage{ragged2e}
\usepackage{wrapfig}
\usepackage{placeins}
\usepackage{needspace}

\usepackage{listings}
\AntTitle{Ask, Condition or Abstain: Reinforcement Learning for Missing-Premise Reasoning}
\AntRunningTitle{Ask, Condition or Abstain}
\AntAuthors{\texorpdfstring{%
  \mbox{Yongqi Tong\textsuperscript{1}\AntEqualContributor} \and
  \mbox{Zhenyu Zhang\textsuperscript{1}\AntEqualContributor} \and
  \mbox{Zimi Liu\textsuperscript{4}} \and
  \mbox{Kewei Fu\textsuperscript{1}} \and
  \mbox{Mingli Song\textsuperscript{2}} \and
  \mbox{Haofei Zhang\textsuperscript{2}} \and
  \mbox{Junshao Zhang\textsuperscript{3}} \and
  \mbox{Hong Zhu\textsuperscript{3}} \and
  \mbox{Jiang-Ming Yang\textsuperscript{1}} \and
  \mbox{Xin Zhang\textsuperscript{1}} \and
  \mbox{Jianshe Li\textsuperscript{1}}%
}{Yongqi Tong, Zhenyu Zhang, Zimi Liu, Kewei Fu, Mingli Song, Haofei Zhang, Junshao Zhang, Hong Zhu, Jiang-Ming Yang, Xin Zhang, Jianshe Li}}
\AntAffiliations{%
  \textsuperscript{1}Ant International\quad
  \textsuperscript{2}Zhejiang University\quad
  \textsuperscript{3}Dingtalk, Alibaba Group\quad
  \textsuperscript{4}Ant Group}
\AntContact{\textbf{Correspondence:} tongyongqi.yq@ant-intl.com}
\AntEqualContributions
\AntDate{\today}
\AntLinks{%
  \href{https://huggingface.co/datasets/ant-intl/MPB_Missing-Premise-Benchmark}{%
    \raisebox{-0.15em}{%
      \includegraphics[height=1.05em]{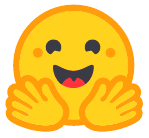}%
    }\hspace{0.3em}Benchmark%
  }%
}
\AntKeywords{reinforcement learning, missing-premise reasoning, abstention, uncertainty, evaluation}
\AntAbstract{%
Answer-only reinforcement learning (RL) trains reasoning models to solve fully specified problems, but many realistic queries omit a premise needed for a unique answer. In this setting, the useful response is not always refusal: the model should ask for the missing premise, condition its answer on the unknown quantity, or abstain when no informative conditional response is available.
We present \emph{Ask-Condition-Abstain Reinforcement Learning} (ACA-RL), a data-augmented RL framework for this setting. Its reasoning-graph-guided pipeline converts well-posed problems into missing-premise training instances with localized gap annotations; ACA-RL then trains on these instances with a structured reward over five observable response behaviors. We also introduce the \emph{Missing-Premise Benchmark} (MPB), a 274-instance human-verified benchmark spanning mathematical, logical, and real-world word problems.
Across Qwen3 and Llama models, ACA-RL consistently improves on MPB while preserving competitive performance on well-posed reasoning tasks.
Together with the released code, MPB, and training data, this work supports a new mission for NLP evaluation: measuring whether models can recognize when a task is underdetermined and handle uncertainty, not only whether they can answer fully specified questions.
}

\begin{document}
\makeanttitle

% \begin{figure}[h!]
%     \centering
%     \includegraphics[width=1\linewidth]{figs/HRBeval2.pdf}
%     \caption{Behavior Score on MPB under the ask/condition/abstain response taxonomy. Strong models often learn to abstain, but still struggle to formulate conditional answers or ask for the missing premise. ACA-RL-trained Qwen3-8B increases these behavior scores relative to generic IDK responses.}
%     \label{fig:mpb_performance}
% \end{figure}

\section{Introduction}

Modern reasoning models are increasingly optimized by reinforcement learning on problems with verifiable final answers. This setting is productive because the reward is clear: solve the problem and match the expected answer. It is also incomplete. A user can omit a rate, leave a relation ambiguous, narrow a definition outside its usable domain, or ask for a value that cannot be identified from the stated premises. In those cases, a model rewarded primarily for definite answers may still produce a confident solution even when the question no longer determines one~\citep{kalai2025languagemodelshallucinate,ouyang2025treecutsyntheticunanswerablemath,sun2024benchmarkinghallucinationlargelanguage}.

The failure mode is not solved by replacing every uncertain response with ``I don't know.'' Abstention is safer than hallucination, but it is often less helpful than explaining the missing variable, giving a conditional expression, or asking the user for the needed premise. We study this response space as a missing-premise reasoning problem: given a query that looks like a standard reasoning problem but lacks information required for a unique answer, the model should decide whether to ask, condition, or abstain instead of fabricating a value.

\begin{figure}[!ht]
    \centering
    \centerline{\includegraphics[width=\linewidth]{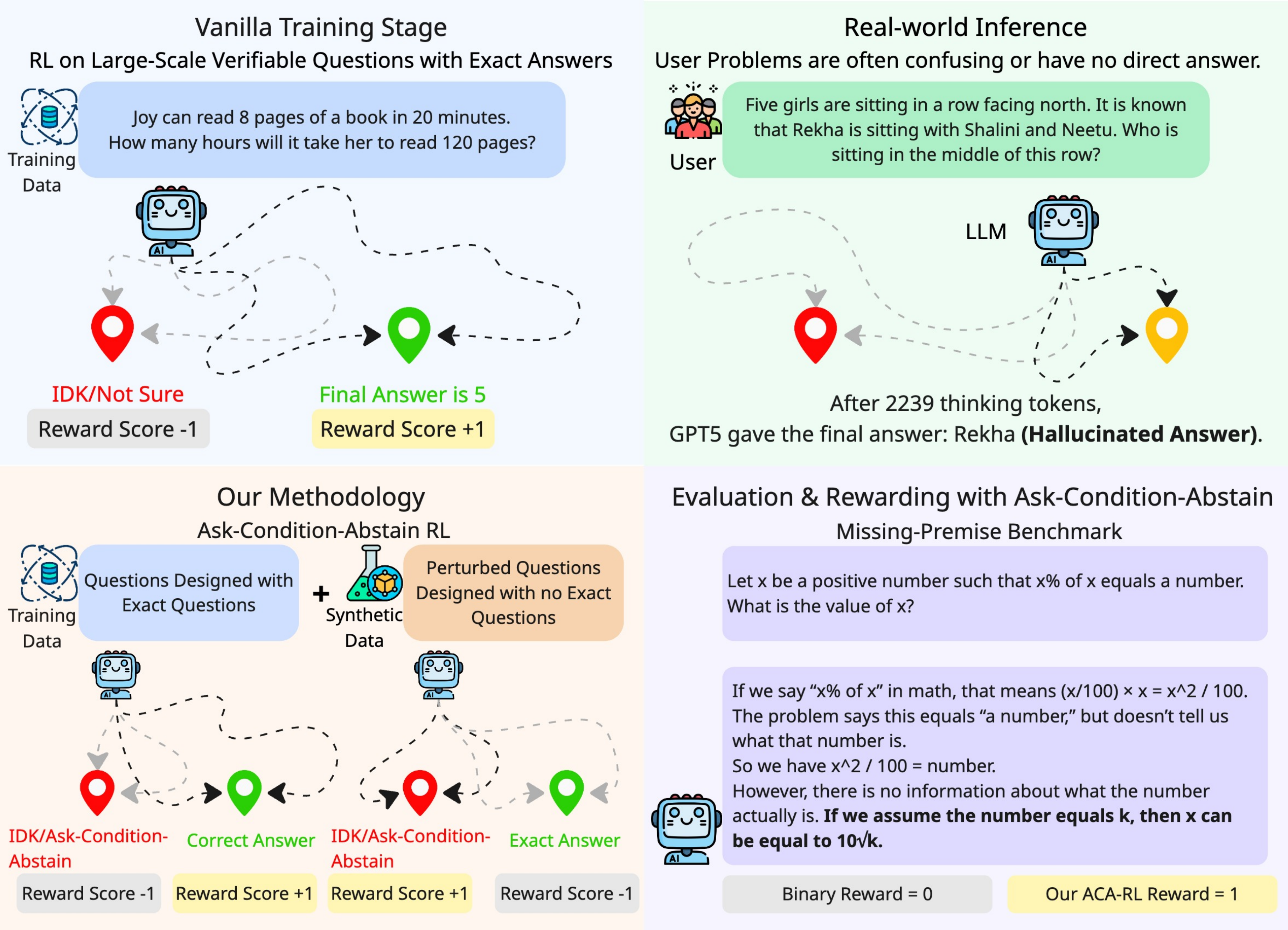}}
    \caption{
    Motivation and framework overview. Answer-only RL trains reasoning models on fully specified questions with verifiable answers, but real user queries often omit critical premises. ACA-RL augments reasoning data with missing-premise variants and uses a structured behavioral reward to prefer asking, conditioning, or abstaining. MPB measures whether responses merely refuse or expose the missing premise in a useful form.}
    \vspace{-6pt}
    \label{fig:overview}
\end{figure}

This framing separates our goal from retrieval and inference-time screening. Retrieval can supply external facts but does not teach the base model to notice underdetermined prompts; screen-then-answer prompting can flag missing premises but depends on an extra inference-time step. We instead learn the missing-premise response policy within the model itself. Our comparison with LM Introspection shows that prompting alone is insufficient for making a model reliably ask, condition, or abstain under missing premises.

We propose \emph{Ask-Condition-Abstain Reinforcement Learning} (ACA-RL), a data-augmented RL framework for learning missing-premise response policies. ACA-RL first uses our reasoning-graph-guided pipeline to decompose well-posed problems, perturb critical premises, and create 120K natural missing-premise training instances with localized gap annotations. It then optimizes a structured reward over five behaviors: silent hallucination, explicit assumption, abstention, conditional formulation, and active elicitation.

\vspace{2pt}
We evaluate this behavior with the \emph{Missing-Premise Benchmark} (MPB), a 274-instance human-verified benchmark spanning six perturbation types across mathematical, logical, and real-world word problems. MPB maps responses to the same five-category taxonomy and reports an averaged Behavior Score, a rubric-aligned behavioral metric rather than a calibrated uncertainty measure.

\vspace{2pt}

Our experiments show that ACA-RL improves MPB and third-party unanswerable benchmark scores over Vanilla PPO, IDK-RL, and LM Introspection while remaining competitive on well-posed reasoning. The gains do not reduce to generic refusal: abstention appears early, while conditional formulation and active elicitation improve more gradually, suggesting a harder response policy beyond IDK. More broadly, the work reframes progress in reasoning models as the ability to recognize when a task is underdetermined and respond constructively, not only as accuracy on fully specified tasks. By teaching models to expose missing premises rather than guess, ACA-RL can support safer reasoning assistants, tutoring systems, and agentic workflows that must decide when to ask users or tools for missing information.

\vspace{3pt}

Our contributions can be summarized as follows:
\vspace{5pt}
\begin{itemize}
\item We formulate missing-premise reasoning as an answer-only RL extension where useful responses ask, condition, or abstain.
\vspace{5pt}
\item We build a reasoning-graph-guided pipeline to synthesize 120K gap-annotated missing-premise instances for ACA-RL training with structured rewards.
\vspace{5pt}
\item We introduce the human-verified MPB benchmark and show that ACA-RL improves missing-premise behavior while preserving competitive standard reasoning performance.
\end{itemize}

% Let page 2 absorb a little more text so the next section does not leave
% a short left column while the right column continues to the page foot.
\section{Preliminary}
\label{sec:preliminary}
A missing-premise problem resembles a well-posed reasoning task but lacks at least one premise needed for a unique answer. Let $s_0$ denote a well-posed source problem and $s'$ its perturbed version with gap annotation $a_{\text{gap}}$. A useful response should not silently fabricate the missing information; it should abstain, formulate the answer conditionally, or ask for the missing premise.

We evaluate terminal response behavior rather than probability calibration. Responses are categorized as silent hallucination, explicit assumption, abstention, conditional formulation, or active elicitation, with ask/condition behaviors preferred to generic refusal and unsupported definite answers penalized. MPB instantiates this taxonomy as a behavioral evaluation protocol: 274 human-verified instances across six perturbation types measure whether a model exposes, parameterizes, or requests missing information rather than merely answering or refusing.

\section{Missing-Premise Data Construction}

ACA-RL requires examples where the original reasoning structure is known and the missing premise can be localized. We build missing-premise data by perturbing well-posed reasoning problems rather than collecting arbitrary ambiguous queries. This treats missing-premise data as a controlled resource for studying robust behavior, not as arbitrary synthetic corruption. The training set contains 120K generated instances, while MPB is constructed from a separate held-out candidate pool and receives additional human expert verification.

\subsection{Reasoning-Graph-Guided Missing-Premise Synthesis}
\label{sec:data}
\begin{figure*}[t]
    \centering
    \centerline{\includegraphics[width=\textwidth]{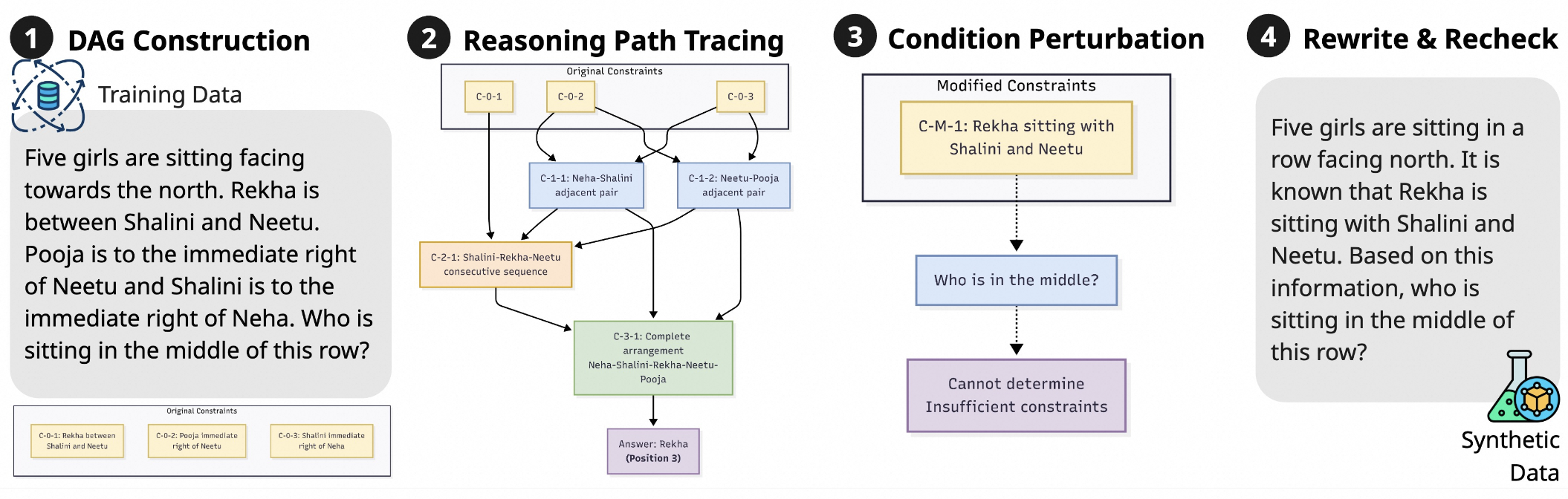}}
    \caption{
    Reasoning-graph-guided missing-premise synthesis pipeline. 
(\textbf{1}) Construct a directed acyclic reasoning graph from a well-posed problem; 
(\textbf{2}) trace the solution path to identify critical constraints; 
(\textbf{3}) perturb key conditions to create a logically underspecified variant; 
(\textbf{4}) rewrite and verify the instance to ensure plausibility and unanswerability. 
The pipeline generates controlled synthetic data for training missing-premise response behavior.
    }
    \label{fig:data_pipeline}
\end{figure*}

To approximate missing-premise cases in a controlled way, we develop a data synthesis pipeline that transforms well-posed problems into plausibly underspecified counterparts. The generated problems contain specific logical gaps rather than random corruption, so the model must detect which premise is missing. The pipeline operates in three stages, and the prompts are shown in Appendix~\ref{sec:datasynthesis}:

\paragraph{Deconstruction and Reasoning Graph Generation.} For a well-posed problem $s_0 \in \mathcal{D}_G$, we first parse it into its constituent components: background, conditions $\{C_0\}$, and question. 
Concurrently, we prompt the model to generate a step-by-step reasoning path, which we structure as a directed acyclic graph (DAG). This graph, or solving tree, makes the dependencies between initial conditions and the final answer explicit.
One detailed example of our DAG is shown in Appendix~\ref{sec:example_dag}.

\paragraph{Surgical Condition Perturbation.}
We then use the reasoning graph to identify a critical condition $c$ on the solution path. A perturbation method is uniformly sampled from our \emph{Conditional Breaking} strategies and applied to $c$, producing a modified condition $c'$. Replacing $c$ with $c'$ in the problem statement, we create an underspecified problem $s'$ with a single, well-defined informational gap. This process yields the pair $(s', a_{\text{gap}})$, where $a_{\text{gap}}$ documents the precise nature of the induced missing premise. Inspired by \citep{sun2024benchmarkinghallucinationlargelanguage}, we formalize the perturbation methods shown in Table~\ref{tab:quant_break_methods}.

\paragraph{Rewrite \& Recheck.} Finally, the perturbed conditions are recomposed with the original background and question into a fluent word problem. The resulting problem $s'$ appears fully specified at first glance, mirroring real-world imperfect information. We use an LLM-based quality filter to assess reasoning-graph correctness and missing-premise unanswerability, discarding samples that fail either check. To validate the unanswerability component, human experts annotate 556 held-out candidate instances; against these labels, the filter reaches 93.0\% accuracy, 94.4\% precision, 91.4\% recall, and 92.9\% F1. We use this filter for scalable training-data construction, while MPB is selected from a separate held-out pool and verified by humans.

This pipeline yields $\mathcal{D}_{\textsc{ACA}}$, a large-scale collection of $(s', a_{\text{gap}})$ pairs in which each $s'$ is a carefully constructed, logically underspecified problem and $a_{\text{gap}}$ documents the nature of its missing premise. By preserving the original reasoning structure while surgically removing or altering critical constraints, the dataset provides controllable scenarios for training and evaluating uncertainty-aware responses. The explicit annotation of informational gaps enables fine-grained reward shaping and benchmarking of missing-premise behavior.

\section{Ask-Condition-Abstain RL}
\label{sec:reward_function}
The desired behavior is broader than simple uncertainty flagging. Instead of only outputting "I don't know" (IDK), a model can ask for the missing premise, condition its answer on an unknown variable, or abstain when neither action is useful. We therefore optimize behavioral robustness under missing premises rather than calibrated probability estimates. The evaluation target is whether the generated response avoids hallucination and communicates the missing premise in a useful form.

ACA-RL operationalizes this response policy with a behavioral reward model that scores both uncertainty detection and localization of the missing premise. Higher rewards are allocated to Conditional Formulation and Active Elicitation, while Abstention remains a positive but lower-valued fallback. The training reward and MPB evaluation use the same behavior taxonomy but separate instances: ACA-RL is optimized on the 120K training set, while MPB is a held-out benchmark for measuring whether the trained policy produces the preferred response categories.

To provide a practical training signal for this objective gap, we design a structured reward function, $R_{\textsc{ACA}}$. Instead of a binary answer/refusal signal, $R_{\textsc{ACA}}$ provides a fine-grained categorical signal across a spectrum of response behaviors. It guides the policy away from unsupported answers and toward more explicit engagement with informational gaps.

\paragraph{A Partition of the Trajectory Space.}
We first partition the space of all possible response trajectories, $\mathcal{T}$, into disjoint sets based on the terminal reasoning behavior exhibited by a trajectory $\tau$. This categorization is performed by a behavior classifier, which implements a classification function, $\text{Behav}(\tau) \to \{\text{SH, EA, Abs, Cond, Elicit}\}$. The behavioral categories are:
\begin{itemize}[itemsep=0pt, parsep=0pt]
    \item \textbf{Silent Hallucination ($\mathcal{T}_{\text{SH}}$):} Trajectories that produce a definite numerical answer by fabricating information without acknowledgment.
    \item \textbf{Explicit Assumption ($\mathcal{T}_{\text{EA}}$):} Trajectories that produce a definite answer but explicitly state the non-grounded assumption made.
    \item \textbf{Abstention ($\mathcal{T}_{\text{Abs}}$):} Trajectories that correctly identify the problem as underspecified and refuse to provide a definite answer.
    \item \textbf{Conditional Formulation ($\mathcal{T}_{\text{Cond}}$):} Trajectories that represent the missing information with a variable and provide a final answer as a formula.
    \item \textbf{Active Elicitation ($\mathcal{T}_{\text{Elicit}}$):} Trajectories that proactively ask a clarifying question to resolve the informational gap.
\end{itemize}
These sets form a partition of the trajectory space: $\mathcal{T} = \mathcal{T}_{\text{SH}} \cup \mathcal{T}_{\text{EA}} \cup \mathcal{T}_{\text{Abs}} \cup \mathcal{T}_{\text{Cond}} \cup \mathcal{T}_{\text{Elicit}}$.

\paragraph{The Reward Value Function.}
We then define a value function, $V: \{\text{SH, EA, Abs, Cond, Elicit}\} \to \mathbb{R}$, that assigns a scalar reward to each behavioral category, reflecting our defined preference hierarchy. The reward for any given trajectory $\tau$ is thus determined by its classification:
\begin{equation}
R_{\textsc{ACA}}(\tau|s') = V(\text{Behav}(\tau))
\end{equation}
The value function $V(b)$ for a behavior $b$ is defined as:

\[
V(b)=
\begin{cases}
1.0 & b=\text{Elicit},\\
0.6 & b=\text{Cond},\\
0.3 & b=\text{Abs},\\
-0.3 & b=\text{EA},\\
-1.0 & b=\text{SH}.
\end{cases}
\]
The reward values encode a preference hierarchy rather than fitted calibration weights:
\[
\text{Elicit} \succ \text{Cond} \succ \text{Abs} \succ \text{EA} \succ \text{SH}.
\]
The acceptable behaviors receive positive rewards, while unsupported assumptions and silent hallucinations receive negative rewards. The highest reward is reserved for active elicitation because it most directly moves the interaction toward acquiring the missing information. Conditional formulation receives the next-highest reward because it makes the unknown variable explicit and preserves useful reasoning without fabricating a value. Abstention remains positive as a safe fallback, but its lower value discourages the model from stopping at generic refusal when it can provide a more informative response.

\paragraph{The ACA-RL Objective.}
With this formal reward structure, we define the ACA-RL objective, $J_{\textsc{ACA}}$, as the expected value over the distribution of underspecified problems under current policy trajectories:
\begingroup
\small
\begin{equation}
J_{\textsc{ACA}}(\theta) = \mathbb{E}_{s' \sim \mathcal{D}_{\textsc{ACA}}} \left[ \mathbb{E}_{\tau \sim \pi_\theta(\cdot|s')} [V(\text{Behav}(\tau))] \right]
\end{equation}
\endgroup
Optimizing $J_{\textsc{ACA}}$ gives the policy an explicit training signal for behaviors that are absent from answer-only supervision. The value function $V(b)$ shifts probability mass away from low-value behaviors such as hallucination ($\mathcal{T}_{\text{SH}}$) and toward higher-value missing-premise responses such as conditional formulation and active elicitation. Thus, $J_{\textsc{ACA}}$ is a practical behavioral proxy for training the model to navigate uncertainty rather than merely replicating fully specified answer paths.

\section{Experiments: Missing-Premise Behavior and General Reasoning}
\label{sec:experiments}

This section evaluates ACA-RL on missing-premise robustness, third-party unanswerable benchmarks, and well-posed reasoning checks. We then study data source, mixture ratio, training steps, and data size to characterize when the method improves the reported Behavior Scores and where it trades off against standard reasoning performance.

\subsection{Experimental Setup}
Details about benchmarks, evaluation methods, and training settings can be found in Appendix~\ref{sec:experiment_details}. We compare ACA-RL against a suite of strong baselines representing different training paradigms:

\begin{itemize}[itemsep=0pt, parsep=0pt]
    \item \textbf{Cold-start SFT}: The cold-start model without any RL fine-tuning. This serves as the base checkpoint for the Qwen experiments. 
    \item \textbf{Vanilla PPO}: A standard verifier-based RL approach~\citep{schulman2017proximalpolicyoptimizationalgorithms} trained via PPO \emph{only} on our set of answerable, well-posed problems, rewarding correct final answers. This baseline represents answer-only RL on fully specified problems.
    \item \textbf{IDK-RL}~\citep{song2025hallucinationtaxreinforcementfinetuning}: A baseline trained to explicitly refuse to answer. It is fine-tuned on a mix of answerable and unanswerable questions, with a binary reward for correctly solving the former and outputting IDK for the latter.
   \item \textbf{ACA-RL (Ours)}: Our proposed framework, trained on a curated set of answerable questions in which approximately 30\% of the instances have been transformed into missing-premise versions via our reasoning-graph-guided condition perturbation pipeline. The structured reward function $R_{\textsc{ACA}}$ defined in Section~\ref{sec:reward_function} explicitly encourages the policy to ask, condition, or abstain on these underspecified problems.
\end{itemize}

\paragraph{Evaluation Protocol.}
The 120K ACA-RL instances are used for training, while MPB is a separate 274-instance benchmark selected from a held-out candidate pool and verified for missing-premise validity. For automatic behavior scoring, we adopt GPT-5 as the judge and map each response to the categories in Section~\ref{sec:reward_function}; Behavior Scores are arithmetic means of the resulting discrete category scores. The same category definitions are used for training-time reward assignment and benchmark scoring, so the results should be read as rubric-aligned behavioral scores on held-out instances. We also report UMWP and SUM as third-party unanswerable benchmarks and standard well-posed reasoning benchmarks to check whether missing-premise training harms ordinary problem solving.

\subsection{Main Results}

Table~\ref{tab:main_results} presents the main results of our experiments. 
As shown in Table~\ref{tab:main_results}, ACA-RL improves missing-premise robustness across all three model groups. On Qwen3-8B, ACA-RL obtains an MPB Behavior Score of 51.73, compared with 8.66 for Vanilla PPO and 48.72 for IDK-RL. The large gap against Vanilla PPO shows that answer-only RL is poorly aligned with missing-premise behavior. The smaller but consistent gap against IDK-RL is the more relevant comparison: IDK-RL learns conservative refusal, while ACA-RL shifts some responses toward conditional formulation and active elicitation under the same rubric.
ACA-RL also keeps general reasoning performance competitive. On Qwen3-8B, the method reduces the drop from Vanilla PPO on GSM8K and AIME'24 compared with IDK-RL, while remaining close on MATH-500. The results therefore support a narrower conclusion: missing-premise training can improve rubric-aligned behavioral robustness while preserving much of the capability learned from well-posed reasoning tasks.

\begin{table*}[htbp]
    \centering
    \tiny
    \caption{Performance across missing-premise and general reasoning benchmarks for different model architectures and scales. Robustness is measured by MPB and by the same ask/condition/abstain Behavior Score on two unanswerable benchmarks (\textbf{UMWP}, \textbf{SUM}); general reasoning is measured by Pass@1 on \textbf{GSM8K}, \textbf{MATH-500}, and \textbf{AIME'24} (averaged over 8 runs). Red values indicate the performance change relative to the strongest RL baseline within each model group.}
    \label{tab:main_results}
    \resizebox{\linewidth}{!}{
    \begin{tabular}{l|c|cc|ccc}
        \toprule
        \multirow{2}{*}{\textbf{Method}} & \textbf{Missing-Premise} & \multicolumn{2}{c|}{\textbf{Unanswerable}} & \multicolumn{3}{c}{\textbf{Mathematics}} \\
        & \textbf{MPB} & \textbf{UMWP} & \textbf{SUM} & \textbf{GSM8K} & \textbf{MATH-500} & \textbf{AIME'24} \\
        \midrule
       \rowcolor[HTML]{F2F2F2} \multicolumn{7}{c}{\textbf{Qwen3-8B (Reasoning Model)}} \\
        Cold-start SFT & 22.81 & 33.53 & 20.51 & 83.69 & 75.20 & 37.08 \\
        Vanilla PPO    & 8.66 & 8.84 & 9.41 & 93.85 & 93.40 & 64.58 \\
        IDK-RL         & 48.72 & 45.74 & 42.95 & 88.02 \color{red}{(-5.83)} & 92.40 \color{red}{(-1.00)} & 63.33 \color{red}{(-1.25)} \\
        ACA-RL (Ours)  & \textbf{51.73} & \textbf{51.50} & \textbf{46.91} & 91.50 \color{red}{(-2.35)} & 92.60 \color{red}{(-0.80)} & 64.16 \color{red}{(-0.42)} \\
        \midrule
        \rowcolor[HTML]{F2F2F2} \multicolumn{7}{c}{\textbf{Qwen3-14B (Reasoning Model) }} \\
        Cold-start SFT  & 29.20 & 43.90 & 28.08 & 88.02 & 87.80 & 43.33 \\
        Vanilla PPO           & 5.29 & 10.38 & 6.16 & 94.84 & 93.60 & 69.17 \\
        IDK-RL    & 49.43 & 52.78 & 41.97 & 89.72 \color{red}{(-5.12)} &  91.20 \color{red}{(-2.40)} & 67.28 \color{red}{(-1.89)} \\
        ACA-RL (Ours)  & \textbf{57.20} & \textbf{56.94} & \textbf{46.56} & 90.60 \color{red}{(-4.24)} & 92.80 \color{red}{(-0.80)} & 69.58 \color{red}{(+0.41)} \\
        
        \midrule
        \rowcolor[HTML]{F2F2F2} \multicolumn{7}{c}{\textbf{Llama3.1-8B (Instruct Model)}} \\
        Instruct       & 3.28 & 12.38 & 5.36 & 80.36 & 46.80 & 7.92 \\
        Vanilla PPO           & 6.18 & 24.73 & 7.13 & 84.15 & 43.20 & 3.33 \\
        IDK-RL    & 51.29 & 54.98 & 44.97 & 83.28 \color{red}{(-0.87)} &  39.00 \color{red}{(-4.20)} & 4.28 \color{red}{(+0.95)} \\
        ACA-RL (Ours)  & \textbf{53.56} & \textbf{55.25} & \textbf{51.58} & 81.43 \color{red}{(-2.72)} & 41.60 \color{red}{(-1.60)} & 4.17 \color{red}{(+0.84)} \\
        
        \bottomrule
    \end{tabular}
    } 
\end{table*}

To further check whether the missing-premise reward induces over-refusal on well-posed tasks, we evaluate Qwen3-8B on LiveBench~\citep{white2024livebench} and SciBench~\citep{wang2023scibench}. Table~\ref{tab:general_capability} shows that ACA-RL matches the Vanilla PPO/RLVR average on LiveBench (59.0 vs. 59.0) and remains competitive on SciBench (53.7 vs. 56.0). This additional evaluation suggests that ACA-RL can improve missing-premise behavior without broadly degrading ordinary problem-solving ability on these benchmarks.

\begin{table}[htbp]
    \centering
    \caption{General capability check on well-posed benchmarks for Qwen3-8B. ACA-RL preserves the LiveBench average and remains competitive on SciBench, indicating that missing-premise training does not collapse into over-refusal on ordinary tasks.}
    \label{tab:general_capability}
    \resizebox{\linewidth}{!}{%
    \begin{tabular}{lcc}
        \toprule
        \textbf{Method} & \textbf{LiveBench Avg.} & \textbf{SciBench Overall} \\
        \midrule
        Cold-start SFT & 51.9 & 47.0 \\
        Vanilla PPO/RLVR & \textbf{59.0} & \textbf{56.0} \\
        ACA-RL (Ours) & \textbf{59.0} & 53.7 \\
        \bottomrule
    \end{tabular}%
    }
\end{table}

\section{Analysis: Data and Behavior Trade-offs}
\label{sec:analysis}
We use the analysis to ask what kind of data and optimization are needed for missing-premise behavior, rather than treating MPB as only another leaderboard. The results distinguish easy-to-learn refusal from more informative ask/condition behavior and expose the trade-off between robustness and standard reasoning.
\paragraph{Data Source for Missing-Premise Behavior.}
Table~\ref{tab:data_source} compares three sources of missing-premise training instances under a fixed data budget: Treecut-style perturbations, SUM-derived unanswerable data, and our reasoning-graph-guided synthesis. Our source obtains the highest MPB Behavior Score in this comparison while maintaining competitive GSM8K and AIME'24 performance, suggesting that the synthesis procedure contributes beyond simply adding unanswerable examples.

\paragraph{Portion of Missing-Premise Questions.}
We investigate how the mixture of answerable and unanswerable problems affects performance. We train variants of ACA-RL with different missing-premise data portions: 10\%, 30\% (our default), 50\%.

The variants in Table~\ref{tab:data_portion} are trained on 10K samples for 100 steps. Results indicate a trade-off between missing-premise robustness and standard reasoning. A higher portion (50\%) gives the highest MPB Behavior Score but lower general reasoning scores, while a lower portion (10\%) behaves more like Vanilla PPO, retaining standard reasoning performance but learning weaker missing-premise behavior. We use 30\% as a practical balance between these two objectives.

\begin{table}[!t]
\centering
\vspace{6pt}
\caption{Data-source ablation under the same data budget. The reasoning-graph-guided synthesis gives the highest MPB Behavior Score while maintaining competitive GSM8K and AIME'24 performance.}
\label{tab:data_source}
    \resizebox{\linewidth}{!}{
    \begin{tabular}{w{l}{1.5cm}|w{c}{1.8cm}|w{c}{1.8cm}w{c}{1.8cm}}
    \toprule
    \textbf{Source} & \textbf{MPB} &  \textbf{GSM8K} & \textbf{AIME'24} \\
    \midrule
    Treecut & 15.41 & 94.01 & 67.50 \\
    SUM     & 47.35 & 91.05 &   59.58\\
    Ours    & 51.73 & 91.50 & 64.16 \\
    \bottomrule
    \end{tabular}
    }
    
\end{table}

\begin{table}[!t]
\centering
\vspace{6pt}
\caption{
    Training on 10K samples over 100 steps shows the trade-off between missing-premise robustness and standard reasoning as the missing-premise ratio changes. We use 30\% as the default because it preserves substantially more general reasoning performance than 50\% while improving MPB over 10\%.
}
\label{tab:data_portion}
    \resizebox{\linewidth}{!}{
    \begin{tabular}{w{l}{1.5cm}|w{c}{1.8cm}|w{c}{1.8cm}w{c}{1.8cm}}
    \toprule
    \textbf{Portion} & \textbf{MPB} &  \textbf{GSM8K} & \textbf{AIME'24} \\
    \midrule
    50\%        & 53.19 & 82.10          & 52.56 \\
    30\%        & 43.79 & 90.14 & 60.83 \\
    10\%        & 28.28          & 92.49 & 63.33 \\
    \bottomrule
    \end{tabular}
    }
\end{table}

\begin{table}[htbp] 
    \centering
    \caption{Comparison between Behavior Score and traditional IDK score. IDK score measures conservative abstention, while Behavior Score additionally rewards responses that use available information or ask for missing premises. ACA-RL improves Behavior Score while retaining competitive IDK behavior.}
    \label{tab:mpb_idk_tradeoff}
    \resizebox{\linewidth}{!}{
    \begin{tabular}{l|cc|cc}
        \toprule
        
        \multirow{2}{*}{\textbf{Method}} & \multicolumn{2}{c|}{\textbf{UMWP}} & \multicolumn{2}{c}{\textbf{SUM}} \\
        & \textbf{Behavior Score} & \textbf{IDK Score}  & \textbf{Behavior Score} & \textbf{IDK Score}  \\
        \midrule
        Cold-start SFT         & 33.53 & 72.57  & 20.51 & 45.77  \\
        Vanilla PPO            & 8.84 & 74.11  & 9.41 & 52.11 \\
        IDK-RL                & 45.74 & \textbf{95.23}  & 42.95 &  \textbf{91.19}\\
        ACA-RL                & \textbf{51.50} & \underline{91.30} & \textbf{46.91} & \underline{84.85}\\
        \bottomrule
    \end{tabular} 
    }
\end{table}

\paragraph{From Abstention to Ask/Condition Behavior.}
The Behavior Score evaluates more behaviors than the IDK score because it also rewards conditional formulation and active elicitation. Table~\ref{tab:mpb_idk_tradeoff} shows that IDK-RL obtains the highest refusal rates, while ACA-RL obtains higher Behavior Scores with slightly lower but still competitive IDK scores. This pattern suggests that ACA-RL shifts some responses from generic refusal toward more informative missing-premise behavior.

\begin{figure}[ht]
    \centering
    \includegraphics[width=\linewidth]{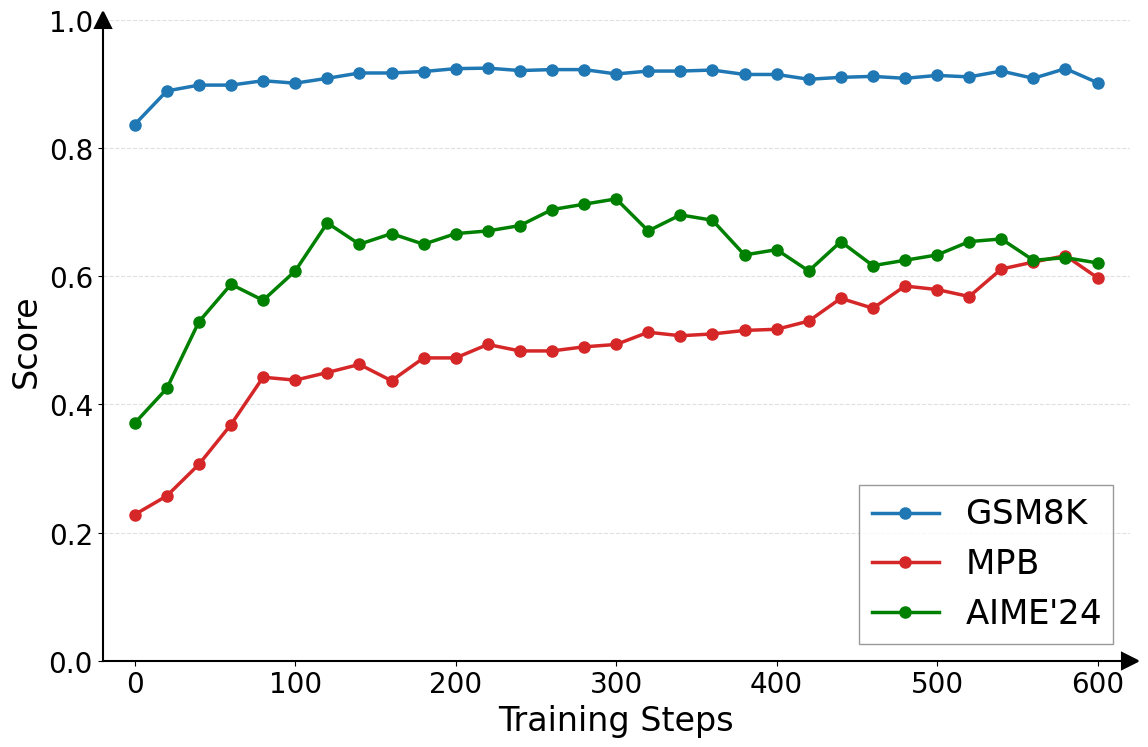}
    \caption{Effect of ACA-RL training steps. Training on our synthetic missing-premise data increases MPB scores in this run while keeping the reported general benchmark scores relatively stable.}
    \label{fig:steps_abl}
\end{figure}

\begin{figure}[ht]
    \centering
    \includegraphics[width=\linewidth]{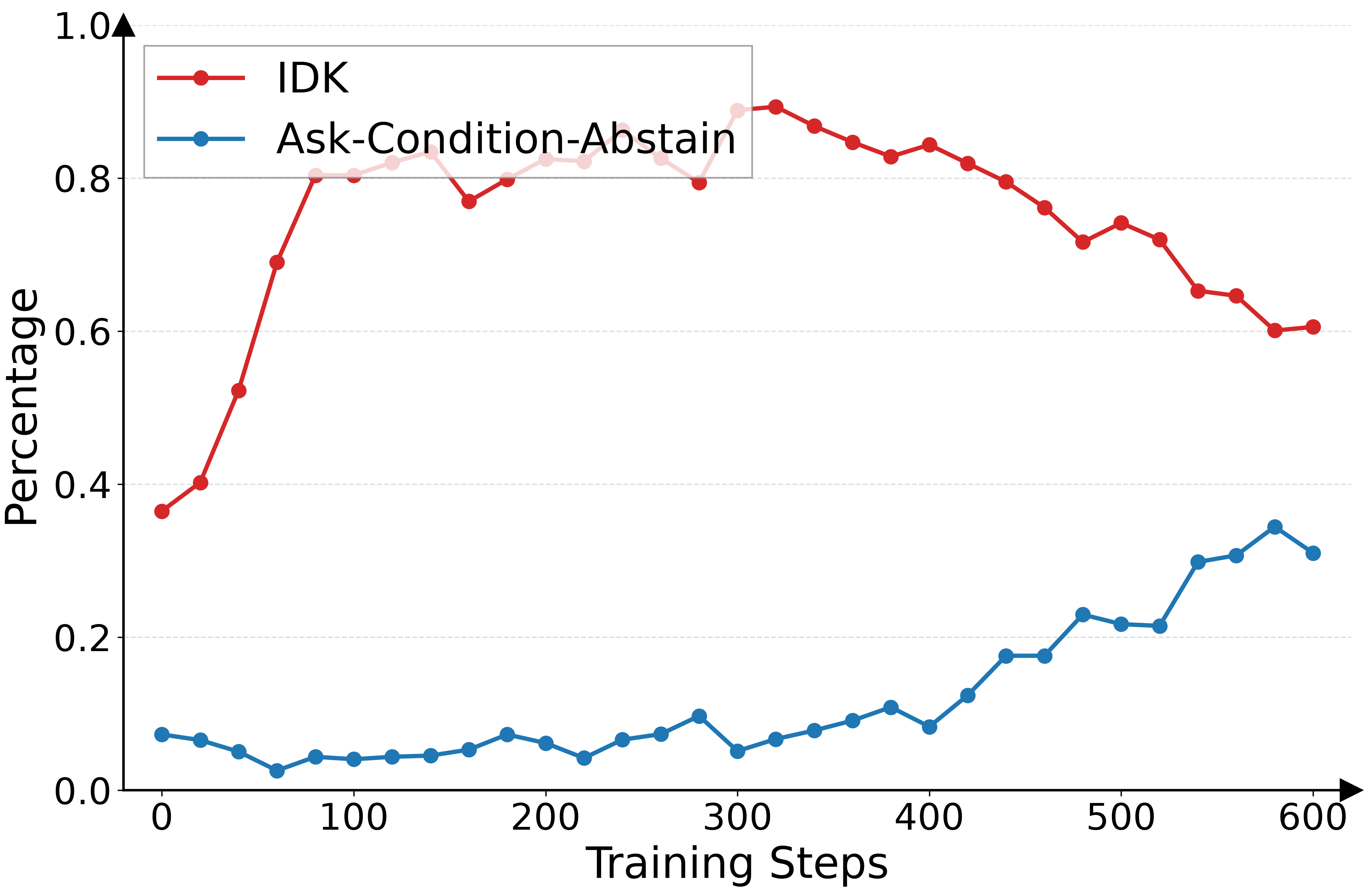}
    \caption{As ACA-RL training progresses, abstention emerges rapidly in early stages, while conditional formulation and elicitation increase more gradually.}
    \label{fig:step_percentage}
\end{figure}

\begin{figure}[!t]
    \centering
    \includegraphics[width=\linewidth]{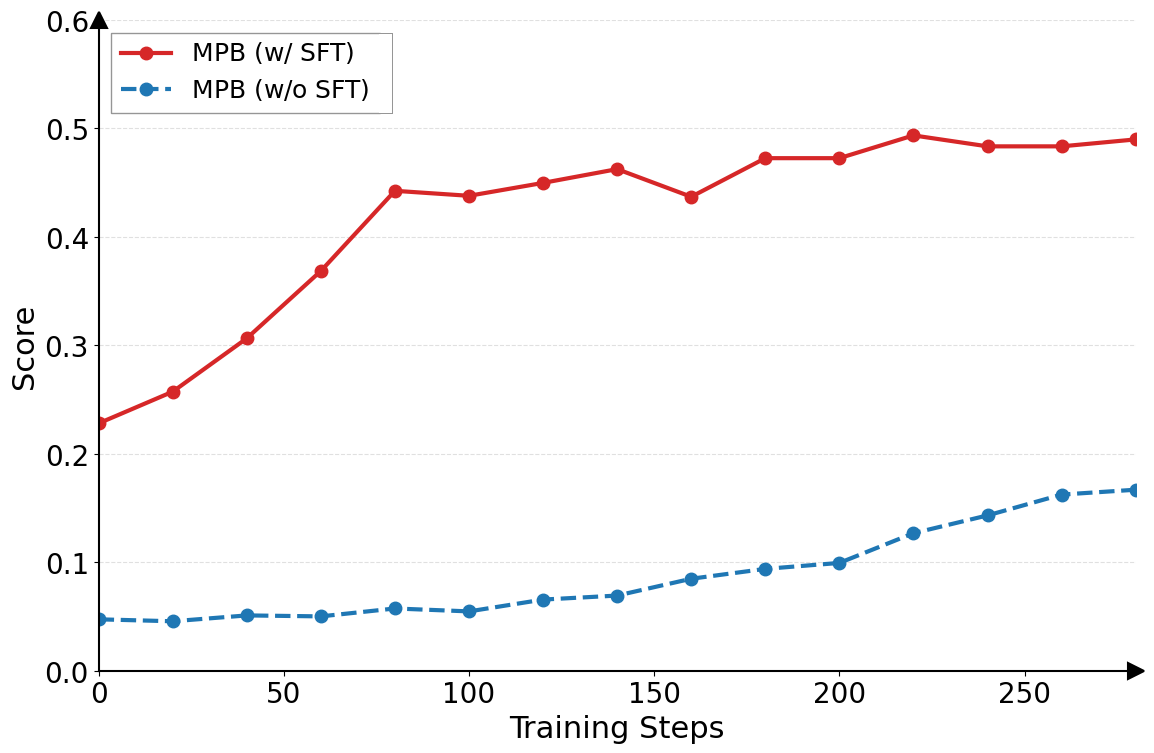}
    \caption{Impact of cold-start SFT on ACA-RL convergence. SFT initialization improves learning efficiency and final MPB score under the plotted training budget.}
    \label{fig:sft_abl}
\end{figure}

\paragraph{Cold-start SFT.}
Figure~\ref{fig:sft_abl} shows that Cold-start SFT improves ACA-RL's learning efficiency on MPB across the training trajectory. Without this initialization, the model learns missing-premise behavior more slowly and reaches a lower final score under the same training budget. This suggests that a supervised warm start provides a useful behavioral prior, while RL is still needed to further refine the ask/condition/abstain policy.

\paragraph{Training Steps.}
As shown in Figure \ref{fig:steps_abl}, while the model's performance on general reasoning benchmarks improves and then plateaus, it exhibits sustained growth on MPB. This suggests that additional training steps can improve the measured missing-premise behavior in this setting while maintaining stable general performance.
Furthermore, Figure \ref{fig:step_percentage} shows that conditional formulation and elicitation gradually displace some IDK responses as training progresses.

\paragraph{Comparison with Training-Free Methods.}
We compare ACA-RL with LM Introspection~\citep{yona2024can}, a training-free method where the model expresses uncertainty through prompting. As shown in Table~\ref{tab:introspection}, ACA-RL obtains a higher MPB score than this prompting baseline (51.73 vs. 9.58). This suggests that explicit missing-premise training is more effective in our setting than prompting alone, though it does not rule out stronger prompting or screen-then-answer systems.

\begin{table}[htbp]
\centering
\caption{Comparison with training-free verbalized uncertainty prompting on Qwen3-8B.}
\label{tab:introspection}
\resizebox{\linewidth}{!}{
\begin{tabular}{w{l}{2.8cm}|w{c}{1.8cm}|w{c}{1.8cm}w{c}{1.8cm}}
\toprule
\textbf{Method} & \textbf{MPB} & \textbf{UMWP} & \textbf{SUM} \\ \midrule
Cold-SFT & 22.81 & 33.53 & 20.51 \\
Vanilla PPO & 	8.66 & 8.84 & 9.41 \\
LM Introspection & 9.58 & 32.01 & 13.99 \\
ACA-RL (Ours) & \textbf{51.73} & \textbf{51.50} & \textbf{46.91} \\ \bottomrule
\end{tabular}
}
\end{table}

\begin{figure}[!t]
    \centering
    \includegraphics[width=\linewidth]{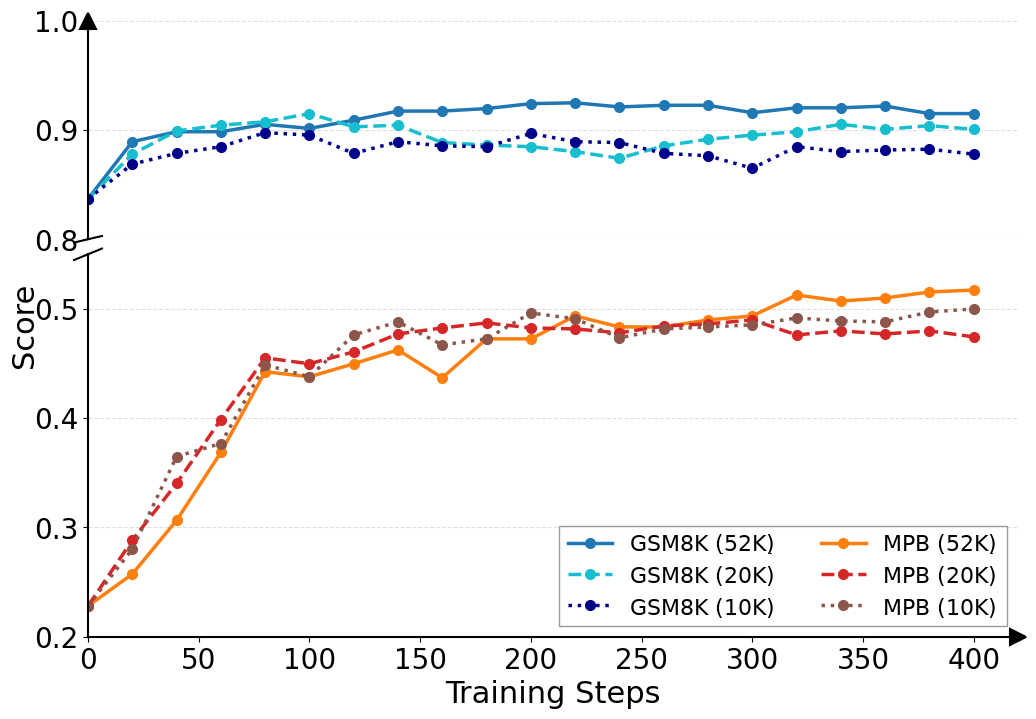}
    \caption{Effect of training data size on ACA-RL performance. Larger generated datasets lead to higher final scores across the plotted training steps on GSM8K and MPB.}
    \label{fig:scaling}
\end{figure}

\paragraph{Training Data Size.}
We investigate the scaling properties of ACA-RL using 10k, 20k, and 52k samples. Figure~\ref{fig:scaling} shows that larger generated datasets improve later-stage performance on both GSM8K and MPB in this sweep, suggesting that additional missing-premise instances provide useful training diversity. This supports the value of scalable missing-premise data construction, though the experiment does not by itself establish a full scaling law.

\section{Related Work}

\paragraph{Reinforcement Learning for Reasoning.}
RL has become a common approach for improving the reasoning behavior of LLMs~\citep{kimiteam2025kimik15scalingreinforcement,deepseekai2025deepseekr1incentivizingreasoningcapability,tong2024optimizinglanguagemodelsreasoning,wang2025bpobalancedpreferenceoptimization,he2025rethinkingreasoningqualitylarge}. Outcome and process rewards provide scalable optimization targets for fully specified tasks~\citep{lambert2024tulu,lightman2023letsverifystepstep,uesato2022solvingmathwordproblems}. Our work focuses on a complementary case: inputs whose premises are missing, ambiguous, or contradictory, where a single final-answer reward is not enough to specify the desired response.

\paragraph{Missing-Premise and Unanswerable Questions.}
Prior work constructs unanswerable or underspecified questions to measure hallucination and abstention behavior. UMWP contains unanswerable MathWorld problems annotated by human experts~\citep{sun2024benchmarkinghallucinationlargelanguage}; Treecut synthesizes unanswerable math problems by removing a dependency edge~\citep{ouyang2025treecutsyntheticunanswerablemath}; and related benchmarks study unreasonable math problems, logical inconsistencies, and abstention failures~\citep{ma2025largelanguagemodelsstruggle,rahman2025blindsolverslogicalthinkers,kirichenko2025abstentionbenchreasoningllmsfail}. MPB extends this line by evaluating a broader behavior taxonomy over missing-premise perturbations rather than only final-answer refusal.

\paragraph{Clarification, Ambiguous QA, and Uncertainty-Aware Reasoning.}
Another line of work studies how LLMs recognize and communicate uncertainty~\citep{tsai2024efficientnonparametricuncertaintyquantification,wang2024llmsknowneedleveraging,huang2025accuracyrolecalibrationselfimproving,ji2025calibratingverbaluncertaintylinear}. Clarification-question work studies when a system should ask a user to resolve ambiguity~\citep{madge2025referentialambiguityclarificationrequests}, while ambiguous-QA settings often emphasize recovering multiple interpretations or plausible answers~\citep{min2020ambigqa}. ACA-RL differs in its training target: asking is only one useful terminal behavior, alongside conditioning on the missing premise and abstaining when no informative conditional answer is available. The method therefore trains a single-turn reasoning policy under missing premises rather than only detecting ambiguity or generating a clarification question.

\paragraph{Abstention and Verbalized Uncertainty.}
Some methods encourage explicit IDK responses~\citep{wu2025lamsslargelanguagemodels}, while others examine symbolic unknowns, prompting-based uncertainty expression, or uncertainty-aware planning~\citep{hu2024uncertaintythoughtsuncertaintyawareplanning,correa2025entropy,li2023counterfactualreasoningtestinglanguage}. ACA-RL is positioned within this family as a reinforcement-learning method that rewards a hierarchy of missing-premise responses: ask, condition, then abstain, with unsupported assumptions and hallucinated answers penalized.

\section{Conclusion and Future Work}
This work argues that reasoning models need a learned response policy for questions whose premises do not determine a unique answer. ACA-RL trains on missing-premise problems generated by a reasoning-graph pipeline and uses structured behavioral rewards to encourage asking, conditioning, or abstaining instead of fabricating a value. Together with MPB, it provides a concrete training and evaluation framework for missing-premise behavior. Across multiple model families, ACA-RL improves Behavior Scores on missing-premise and unanswerable benchmarks while preserving competitive performance on well-posed reasoning tasks in the reported evaluations. This points toward a broader mission for NLP systems: models should not only answer when information is complete, but also recognize when interaction, retrieval, or tool use is required to make a task answerable.

\section*{Limitations}
Although ACA-RL trains models to detect missing premises, Active Elicitation is evaluated as a single-turn textual response. The method does not train retrieval, tool use, or multi-turn clarification policies that can actually acquire the missing premise. This leaves a gap between identifying an underspecified problem and resolving it in agentic workflows.

Furthermore, our training data is derived from a structural perturbation pipeline applied to logical problems. While efficient, these synthetic "broken" links may not fully capture the messy, implicit, or semantic ambiguity found in organic user queries. As a result, the model's robustness is primarily verified against logical incompleteness rather than the full spectrum of open-world ambiguity.

Finally, both training-time reward assignment and MPB scoring use the same behavior taxonomy. The held-out MPB split prevents instance-level training leakage, but future work should still test independent scoring protocols and conduct larger-scale human scoring. The margins over IDK-RL should therefore be interpreted as gains under the reported Behavior Score rubric.

\bibliographystyle{plainnat}
\bibliography{references}
\clearpage
\appendix
\onecolumn

\section{MPB: Construction and Diagnostics}
\label{sec:benchmark}

\noindent
\begin{minipage}[t]{0.48\textwidth}
\vspace{0pt}
Current reasoning benchmarks are largely confined to well-posed problems and therefore do not directly assess model behavior under imperfect information. 
Related work on unanswerable math questions often measures hallucination or abstention rates, rather than the broader response categories studied here.
To address this evaluation gap, we introduce the \emph{Missing-Premise Benchmark} (MPB), a benchmark for measuring model behavior when confronted with underspecified or inconsistent problem statements.
MPB comprises 274 well-curated instances spanning mathematical, logical, and real-world word problems, each intentionally designed with missing or conflicting premises.
Source problems are drawn from \citet{tong2023eliminatingreasoninginferringplanning} and DeepscaleR~\citep{deepscaler2025}, then transformed by the missing-premise synthesis pipeline before filtering and verification.

The MPB benchmark is curated from a held-out candidate pool generated by our pipeline (Section~\ref{sec:data}) through a multi-stage filtering process and is not included in the 120K ACA-RL training set. 
The process begins with automated checks to verify the successful perturbation of each problem and filter out malformed outputs. Surviving candidates then undergo a two-tier qualitative review: first, an automated quality filter screens each problem for naturalness, plausibility, and subtlety; then, three human experts, each with a graduate degree, verify the problem pair's validity ($s_0, s'$), the analysis's accuracy ($a_{\text{gap}}$), and overall quality.
\end{minipage}
\hfill
\begin{minipage}[t]{0.48\textwidth}
\vspace{0pt}

Evaluation on MPB classifies each response using our behavioral hierarchy (Section~\ref{sec:reward_function}) to yield a distribution of response behaviors. This diagnostic view separates silent hallucination, explicit assumption, abstention, conditional formulation, and active elicitation.
Detailed grading instructions are provided in Appendix~\ref{sec:llmjudge}.

The results in Table~\ref{tab:res} suggest that MPB difficulty varies across perturbation types. Numerical value removal and relationship unquantifiable replacement receive higher scores for several models, while condition contraction and qualifier disruption remain difficult. The two inference modes, \emph{Thinking} and \emph{Direct}, also show different behavior patterns. No listed model is uniformly strong across all six perturbation categories, which supports using MPB as a diagnostic benchmark for missing-premise behavior.
\end{minipage}

\vspace{10pt}
\begin{center}
\captionsetup{hypcap=false}
\captionof{table}{Diagnostic MPB Behavior Scores across six missing-premise perturbation tasks.
Abbreviations: Rel.r = Relationship removal; Rel.unquan.r = Relationship unquantifiable replacement; Num.val.r = Numerical value removal; Enti.dis = Entity disruption; Qual.dis = Qualifier disruption; Cond.con = Condition contraction. The final column reports the benchmark-level ask/condition/abstain Behavior Score.}
\label{tab:res}
\resizebox{\textwidth}{!}{
\begin{tabular}{l c c c c c c c c}
\toprule
\textbf{Models} & \textbf{Reasoning Mode} & \textbf{Rel.r} & \textbf{Rel.unquan.r} & \textbf{Num.val.r} & \textbf{Enti.dis} & \textbf{Qual.dis} & \textbf{Cond.con} & \textbf{MPB score} \\
\midrule
\multirow{2}{*}{Qwen3-235B-A22B-Instruct}
& Thinking & 5.29 & 12.45 & 4.61 & 1.96 & 0.36 & 0.18 & 6.75 \\
& Direct & 6.48 & 10.68 & 4.33 & 2.01 & 0.41 & 0.18 & 12.32 \\
\multirow{2}{*}{Qwen-Plus}
& Thinking & 5.79 & 10.86 & 4.33 & 2.24 & 0.59 & 0.09 & 7.48 \\
& Direct & 5.16 & 13.14 & 5.52 & 2.10 & 0.68 & 0.09 & 6.66 \\
Qwen3-Next-80b-A3B & Thinking & 5.66 & 12.32 & 4.88 & 2.46 & 0.36 & 0.18 & 7.12 \\
Deepseek-V3.1 & Direct & 6.61 & 11.54 & 5.16 & 2.10 & 0.78 & 0.00 & 5.57 \\
Qwen2.5-72b-Instruct & Direct & 6.34 & 11.59 & 4.84 & 2.14 & 0.27 & 0.05 & 8.03 \\
Deepseek-R1 & Thinking & 6.57 & 12.91 & 4.61 & 1.92 & 0.46 & 0.00 & 6.20 \\
Llama-3.3-70B-Instruct & Direct & 6.02 & 12.09 & 5.47 & 1.87 & 0.46 & 0.09 & 4.11 \\
QwQ-32B & Thinking & 6.43 & 11.18 & 4.74 & 1.87 & 0.46 & 0.00 & 9.76 \\
GPT-5 & -- & 6.43 & 11.50 & 5.06 & 2.28 & 0.41 & 0.09 & \textbf{20.07} \\
\bottomrule
\end{tabular}
}
\end{center}

\clearpage

\section{Condition Perturbation Definitions and Examples}
Detailed definitions and examples of the missing-premise construction methods are shown in Table~\ref{tab:quant_break_methods}.

\begin{center}
\captionsetup{hypcap=false}
\captionof{table}{Condition-perturbation operations used to construct missing-premise instances, with examples from MPB. Each \textbf{Perturbed Question} lacks enough information for a direct answer, yet a strong model may still produce a confident response.}
\label{tab:quant_break_methods}
\vspace{4pt}
\begingroup
\tiny
\resizebox{\textwidth}{!}{
\begin{tabular}{p{3cm}|p{3cm}|p{3cm}|p{3cm}}
\toprule
\textbf{Construction Methods} & \textbf{Original Question}  & \textbf{Perturbed Question} & \textbf{Example Model Response}\\
\midrule
\textbf{Relationship Removal} - If the condition involves a relationship between two entities, remove that relationship. & Five girls are sitting facing towards the north. Rekha is between Shalini and Neetu. Pooja is to the immediate right of Neetu, and Shalini is to the immediate right of Neha. Who is sitting in the middle? & Five girls are sitting in a row facing north. It is known that Rekha is sitting with Shalini and Neetu. Based on this information, who is sitting in the middle? & The response treats "sitting with" as enough ordering information and returns a definite middle person, although the missing relation makes the arrangement underdetermined.\\
\midrule
\textbf{Relationship Unquantifiable Replacement} - Transform a definite quantitative relationship into an indefinite or non-numeric one. & Jayant introduces a man to his friend as his wife's father's son. The man is Jayant's...? & While attending a family gathering, Jayant introduces a man to his cousin, explaining that the man is part of his wife's father's extended family. Jayant's cousin, curious about their relationship, tries to determine the exact connection. The man is Jayant's ...? & The response gives a broad in-law relation but still presents it as a resolved answer even though the exact family relation is no longer specified.\\
\midrule
\textbf{Numerical Value Removal} - Directly remove the numerical value from the quantitative relationship. & In a mathematics contest with ten problems, a student gains 5 points for a correct answer and loses 2 points for an incorrect answer. If Olivia answered every problem and her score was 29, how many correct answers did she have? & In a mathematics contest consisting of ten problems, each correct answer adds points to a contestant's score, while each incorrect answer subtracts points. Olivia attempted every problem, and after the contest, her total score was 29. How many of the ten problems did Olivia answer correctly? & The response invents point values and returns a numeric answer, even though the scoring rule has been removed.\\
\midrule
\textbf{Qualifier Removal} - Remove the qualifier. & Extend the square pattern of 8 black and 17 white square tiles by attaching a border of black tiles around the square. What is the ratio of black tiles to white tiles in the extended pattern? & In a tile pattern made up of black and white square tiles arranged in a square, a border of tiles is attached around the square. What is the ratio of black tiles to white tiles in the extended pattern? The answer should be a numeric value. & The response assumes a particular square size and border configuration, then returns a numeric ratio despite the missing tile counts.\\
\midrule
\textbf{Qualifier Disruption} - Replace the qualifier with a specific but contextually irrelevant condition, thereby invalidating the original constraint while ensuring the new condition remains plausible and non-absurd. & There are 6 people in a family. Each one of them likes a different colour: Blue, Red, Pink, Green, Yellow, and White. Seema, who likes Red, is Anitha's mother-in-law and Anitha is Raja's wife. Dinesh is Rohan's father who dislike the blue or white. Bavya likes the yellow and is Rohan's sister, who likes Pink. Raja does not use White. Which colours is liked by Anitha? & In a family where each member likes a different color and has various relationships to one another, there is one known preference: Raja does not use White colour when it is raining. Given this information and the possible color choices of red, blue, green, yellow, and white, which colours is liked by Anitha? & The response solves a full logic-grid puzzle even though most family and color constraints have been removed.\\
\midrule
\textbf{Entity Disruption} - Replace the condition's entity with a different but contextually plausible entity that is unrelated to the problem, ensuring it does not resemble or ambiguously refer to any existing entity and the resulting sentence remains grammatically correct and natural. & For how many ordered pairs (b,c) of positive integers does neither $x^2+bx+c=0$ nor $x^2+cx+b=0$ have two distinct real solutions? & Let b and c be positive integers. Consider the two quadratic equations $x^2 + bx + z = 0$ and $x^2 + zx + b = 0$. It is given that neither of these quadratics has two distinct real solutions. How many ordered pairs (b, c) of positive integers satisfy this condition? & The response treats the unrelated variable $z$ as if it were specified and counts ordered pairs for $b,c$ anyway.\\
\midrule
\textbf{Condition Contraction} - Narrow a broadly applicable condition into a more specific one, thereby causing partial information loss and breaking the original coverage. & For any positive integer n, define [n] to be the sum of the positive factors of n. For example, [6] = 1 + 2 + 3 + 6 = 12. Find [[11]]. & For any positive integer n less than 10, [n] is defined as the sum of the positive factors of n, and it is given that [6] equals 12. Find [[11]]. & The response applies the definition outside its stated domain and returns the original well-posed answer.\\
\bottomrule
\end{tabular}}
\endgroup
\end{center}

\clearpage
\section{Reasoning-Graph Example}
\label{sec:example_dag}
An example reasoning graph is shown in Figure~\ref{fig:dag_example}.

\begin{center}
\captionsetup{hypcap=false}
\includegraphics[width=0.62\textwidth]{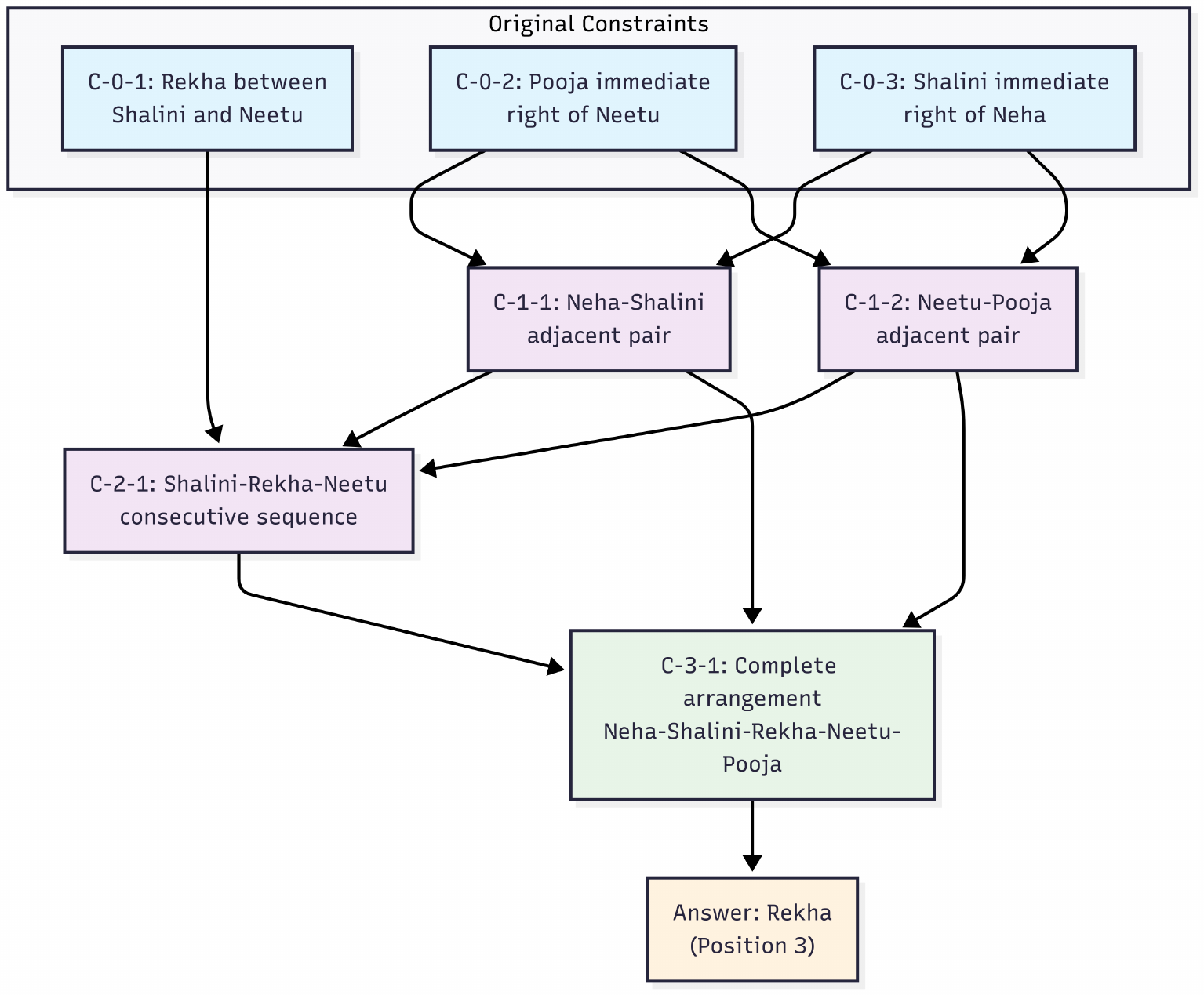}
\captionof{figure}{An example of our reasoning DAG illustrating how original constraints are progressively combined into intermediate inferences, ultimately yielding the complete arrangement and final answer.}
\label{fig:dag_example}
\end{center}

\clearpage
\twocolumn
\section{Behavioral Scoring Example}
\label{sec:example_eval}

\begin{tcolorbox}[
    fontupper=\small,
    colback=white,
    colframe=black,
    coltitle=white,
    fonttitle=\bfseries,
    title=Scoring Example,
    colbacktitle=gray,
    boxrule=0.5pt,
    arc=5pt,
    width=\linewidth,
    toptitle=0.2mm,
    bottomtitle=0.2mm
]
\textbf{Question.} Two parabolic curves with constants $a$ and $b$ intersect the coordinate axes, producing four intercept points. These points form a kite whose area is given. What is $a+b$?

\textbf{Missing premise.} The problem does not specify the actual equations of the parabolas or enough constraints to determine $a$ and $b$.

\textbf{Undesired response.} A silent-hallucination response chooses a symmetric configuration, assigns unstated equations, and returns a numeric value.

\textbf{Preferred response.} A conditional-formulation response states that $a+b$ is not identifiable from the given information and either expresses the answer in terms of additional parameters or asks for the missing equations/constraints.
\end{tcolorbox}

\section{Experiment Details}

\subsection{Training and Evaluation Details}
\label{sec:experiment_details}

\paragraph{Base Model and Implementation.}
To evaluate ACA-RL across different architectures and scales, we use two series of models: the \textbf{Qwen3} family (8B and 14B)~\citep{yang2025qwen3} as representative reasoning-heavy models, and \textbf{Llama3.1-8B-Instruct}~\citep{grattafiori2024llama} as a representative instruction-tuned model. 
For the Qwen3 series, we initially perform a supervised fine-tuning (SFT) cold-start on our synthesized data, yielding the \textit{Cold-start SFT} checkpoints as the initialization for subsequent RL phases. In contrast, for Llama3.1-8B-Instruct, we directly use the vanilla instruct model as the baseline without any additional SFT distillation of "Chain-of-Thought" (CoT) or "Think" trajectories. This preserves the model's original instruction-following setup rather than adding an extra reasoning-trace distillation stage.
All reinforcement learning experiments are conducted using the PPO algorithm~\citep{schulman2017proximalpolicyoptimizationalgorithms} implemented via the open-source veRL library~\citep{sheng2025hybridflow}. The KL coefficient is set to 0.0 to allow broad policy exploration under missing-premise environments. We keep the total number of answerable training queries and computational budget consistent across all RL baselines, and the synthetic data is identical between IDK-RL and ACA-RL. Detailed hyperparameters, prompt templates, and training configurations are provided in Table~\ref{tab:ppo_hyperparams}. In the ablation analysis presented in Section \ref{sec:analysis}, we use Qwen3-8B as a representative model.

\paragraph{Evaluation Datasets.}
We evaluate all models on two categories of benchmarks:
\begin{itemize}[itemsep=0pt, parsep=0pt]
    \item \textbf{Benchmarks}: Our proposed \textbf{MPB} benchmark, along with two other unanswerable question datasets, UMWP~\citep{sun2024benchmarkinghallucinationlargelanguage} and SUM~\citep{song2025hallucinationtaxreinforcementfinetuning}, to assess out-of-distribution (OOD) robustness.
    \item \textbf{General Reasoning Benchmarks}: High-difficulty, well-posed math and logic benchmarks, including AIME'24~\citep{aime2024}, MATH-500~\citep{hendrycksmath2021}, and GSM8K~\citep{cobbe2021gsm8k}, to measure general reasoning capabilities.
\end{itemize}
For MPB, we report Behavior Score and IDK score (\S\ref{sec:reward_function}). For UMWP and SUM, we report the same Behavior Score alongside IDK score when used in the analysis (Section~\ref{sec:llmjudge}). For general reasoning benchmarks, we report Pass@1 on GSM8K and MATH-500, and average Pass@1 over 8 samples on AIME'24.

\paragraph{Data Source.} 
Our missing-premise training data is synthesized exclusively from DeepscaleR \citep{deepscaler2025}, where the answerable questions are taken directly from the original dataset for training.

\subsection{MPB Scoring}
\label{sec:mpb_scoring}
For the MPB benchmark, we use a GPT judge. Each response is first classified into one of the five behavior categories in Section~\ref{sec:reward_function}. The category is then mapped to a discrete Behavior Score using Table~\ref{tab:score_mapping}, and benchmark-level values are arithmetic means over test instances. This is why MPB results appear as continuous values in the main tables even though each individual response receives one of five discrete scores.

\begin{table}[ht]
\centering
\caption{Mapping between ask/condition/abstain behavior labels and final Behavior Scores.}
\label{tab:score_mapping}
\resizebox{0.48\textwidth}{!}{ 
\begin{tabular}{ccl}
\toprule
\textbf{Reward Score} & \textbf{Behavior Score} & \textbf{Description} \\
\midrule
-1.0 & 0 & Silent Hallucination \\
-0.3 & 25 & Explicit Assumption \\
 0.3 & 50 & Abstains from Answering \\
 0.6 & 75 & Conditions on a Variable \\
 1.0 & 100 & Asks for the Missing Premise  \\
\bottomrule
\end{tabular}
}
\end{table}

\subsection{Binary IDK Scoring}
\label{sec:idk}
The IDK scoring process is based on the same behavioral framework but is adapted to a binary rubric (0 for incorrect, 1 for correct). In line with the evaluation protocol of \citet{song2025hallucinationtaxreinforcementfinetuning}, we append the instruction ``If you don't know the answer, reply with \verb|\boxed{I don't know.}|'' to each question prompt.

\begin{table}[t]
        \centering
        \caption{Hyperparameters of the PPO algorithm implemented based on the veRL framework.}
        \label{tab:ppo_hyperparams}
        
        \resizebox{0.95\linewidth}{!}{ 
        \begin{tabular}{w{c}{3.5cm}w{c}{5cm}w{c}{5cm}}
        \toprule
             \textbf{Category} &  \textbf{Hyperparameter} & \textbf{Value} \\ 
             \midrule
             \multirow{4}{*}{Trainer} &  Nodes & 4\\
                                       &  GPUs per node & 8\\
                                       &  Total steps & 400\\
                                       &  Gradient checkpointing & True \\
             \midrule
             \multirow{2}{*}{Algorithm} &  Advantage estimator & GAE($\lambda$=1, $\gamma$=1) \\
                                        &  Use KL in reward & False\\
             \midrule
             \multirow{6}{*}{Actor} &  Learning rate & $1\times10^{-6}$ \\
                                     &  Mini-batch size & 128 \\ 
                                     &  Clip ratio & 0.2 \\ 
                                     &  Entropy coefficient & 0 \\ 
                                     &  Use dynamic batch size & True \\ 
                                     &  Ulysses sequence parallel size & 4 \\ 
             \midrule
             \multirow{4}{*}{Rollout} &  Backend & vLLM \\
                                       &  Temperature  & 1.0 \\ 
                                       &  Top-p  & 1.0 \\ 
                                       &  Tensor model parallel size & 2 \\ 
             \midrule
             \multirow{3}{*}{Critic} &  Learning rate & $1\times10^{-6}$ \\
                                      &  Warm-up steps & 0 \\ 
                                      &  Ulysses sequence parallel size & 4 \\ 
             \midrule
             \multirow{2}{*}{Reward Judge} &  Judge type & GPT-based judge \\ 
                                            &  Output & Five behavior categories \\ 
             \midrule
             \multirow{2}{*}{Data} &  Batch size & 512 \\
                                    &  Max response length & 14000 \\ 
             \bottomrule
        \end{tabular}
        }
\end{table}

\subsection{Additional Evaluator Agreement Check}
\label{sec:llmjudge}
\paragraph{Setting.}
For MPB scoring, we adopt GPT-5 as the judge. Each response is assigned to one terminal behavior category, and the category is then mapped to the reward or Behavior Score shown in Table~\ref{tab:score_mapping}. As an additional reliability check, three human experts, each with a graduate degree, annotated sampled GPT-5 and Qwen3-235B-A22B-Instruct responses using the same behavior labels.

\paragraph{Conclusion.}
Human labels agree with the GPT judge on approximately 98\% of the sampled GPT-5 responses and 95\% of the sampled Qwen3-235B-A22B-Instruct responses. This agreement check supports the consistency of the reported MPB scoring protocol and broader evaluator independence.

\section{Data Synthesis Details}
\label{sec:datasynthesis}
This process is implemented as an LLM-agent workflow. For reproducibility, the released supplement will include the prompt templates used for problem decomposition, reasoning graph generation, surgical condition perturbation, reconstruction, and rechecking. We summarize the role of each stage below rather than embedding the raw prompt files in the paper body.

\subsection{Problem Decomposition}
The decomposition prompt asks the model to separate each source problem into background information, explicit conditions, and the target query. This representation provides the input structure for reasoning-graph construction.

\subsection{Reasoning Graph Generation}
The reasoning-graph prompt asks the model to convert a solution into a directed acyclic graph whose nodes are conditions or intermediate conclusions and edges represent dependency relations.

\subsection{Surgical Condition Perturbation}
The perturbation prompt selects a condition on the reasoning path and applies one of the condition-breaking operations in Table~\ref{tab:quant_break_methods}. The goal is to remove or alter a premise that is necessary for a unique answer while keeping the edited question fluent and plausible to a reader.

\Needspace{6\baselineskip}
\subsection{Reconstruction}
The reconstruction prompt rewrites the perturbed conditions, original background, and target query into a fluent problem statement.

\subsection{Recheck}
\subsubsection{Reasoning Correctness}
The reasoning-correctness check verifies that the original reasoning graph and answer remain coherent before perturbation.

\subsubsection{Missing-Premise Unanswerability}
The missing-premise check verifies that the perturbed problem no longer contains enough information to determine the original answer and that the induced gap is identifiable.

\section{Use of LLMs}
LLMs are used as methodological components in this work for data synthesis, filtering, and GPT-based behavioral judging, as described in the method and experiment sections. Separately, general-purpose LLM writing assistants may have been used for minor wording, grammar, and formatting support. The research questions, method design, experimental analysis, and final claims are the responsibility of the authors.

\end{document}